\documentclass{article}
\usepackage{amssymb}
\usepackage{amsmath}

\usepackage[final]{corl_2025}
\usepackage{graphicx}
\usepackage{wrapfig}
\usepackage{enumitem}

\title{GelFusion: Enhancing Robotic Manipulation under Visual Constraints via Visuotactile Fusion}

\author{
    Shulong Jiang\textsuperscript{1} ~~ Shiqi Zhao\textsuperscript{1} ~~ Yuxuan Fan\textsuperscript{2} ~~ Peng Yin\textsuperscript{1} \\[2mm]
    \normalsize \textsuperscript{1}City University of Hong Kong \\
    \vspace{2mm}
    \textsuperscript{2}The Hong Kong University of Science and Technology (Guangzhou) \\
    \textup{\url{https://gelfusion.github.io/}}
    \vspace{-5mm}
}

\begin{document}
\maketitle

\begin{abstract}
Visuotactile sensing offers rich contact information that can help mitigate performance bottlenecks in imitation learning, particularly under vision-limited conditions, such as ambiguous visual cues or occlusions. Effectively fusing visual and visuotactile modalities, however, presents ongoing challenges.
We introduce GelFusion, a framework designed to enhance policies by integrating visuotactile feedback, specifically from high-resolution GelSight sensors. GelFusion using a vision-dominated cross-attention fusion mechanism incorporates visuotactile information into policy learning. To better provide rich contact information, the framework's core component is our dual-channel visuotactile feature representation, simultaneously leveraging both texture-geometric and dynamic interaction features.
We evaluated GelFusion on three contact-rich tasks: surface wiping, peg insertion, and fragile object pick-and-place. Outperforming baselines, GelFusion shows the value of its structure in improving the success rate of policy learning.
\end{abstract}


\begin{figure}[h]
    \centering
    \includegraphics[width=\textwidth]{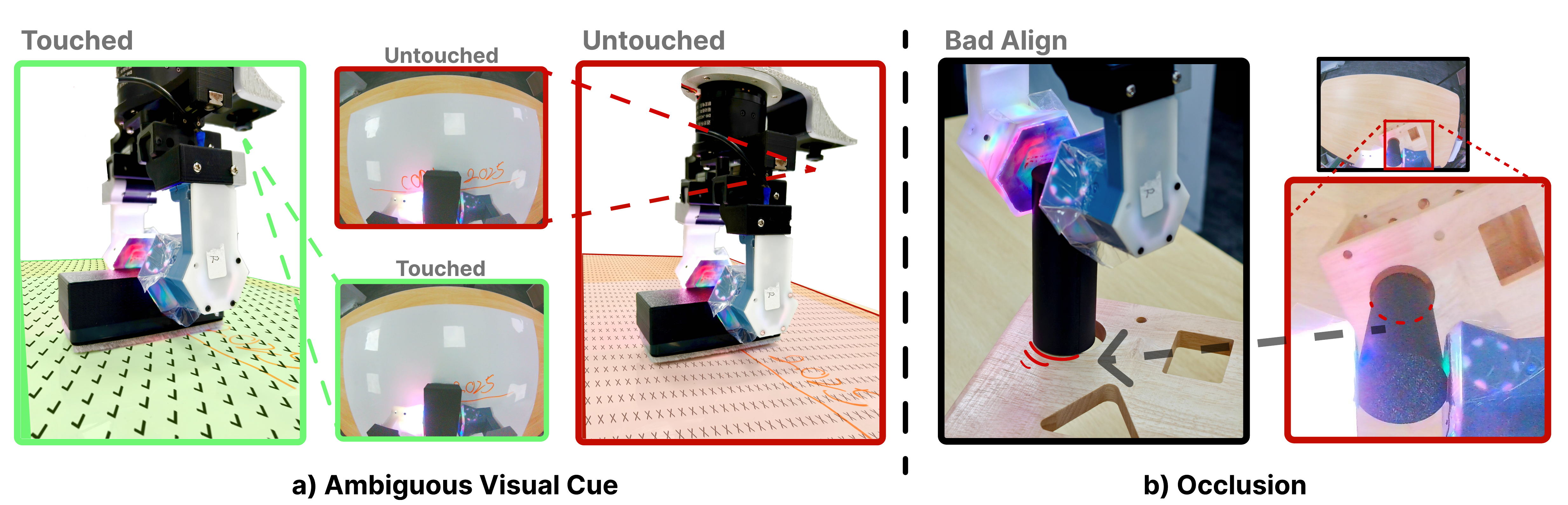}
    \vspace{-5mm}
    \caption{\textbf{Visionlimited conditions for policy learning. } More specifically,
(a) ambiguous visual cues specifically refers to situations where, when the field of view is restricted, it is difficult to obtain complete state estimation based solely on visual information. An example is the scene degradation that occurs when trying to erase a whiteboard, making depth difficult to estimate.
(b) occlusion frequently occurs in operations involving assembly or tool use, often when the gripped object blocks the details required for critical steps.}
    \label{fig:teaser}
\end{figure}
\vspace{-2mm}

\section{Introduction}
\vspace{-2mm}
\label{sec:intro}

The performance of a good Imitation Learning policy strongly depends on complete observations. However, prevailing vision-based systems often struggle in contact-rich scenarios due to limitations like ambiguous visual cues or occlusions(Fig. \ref{fig:teaser}). High-resolution visuotactile sensors, such as GelSight \cite{yuan2017gelsight}, offer a promising avenue by providing rich contact information detailing surface properties and interaction dynamics, often inaccessible through vision alone. Despite this potential, effectively fusing visual and visuotactile modalities for robust policy learning remains a significant challenge, risking information loss or dominance by one modality if not carefully designed.

\begin{figure}
    \centering
    \includegraphics[width=0.99\textwidth]{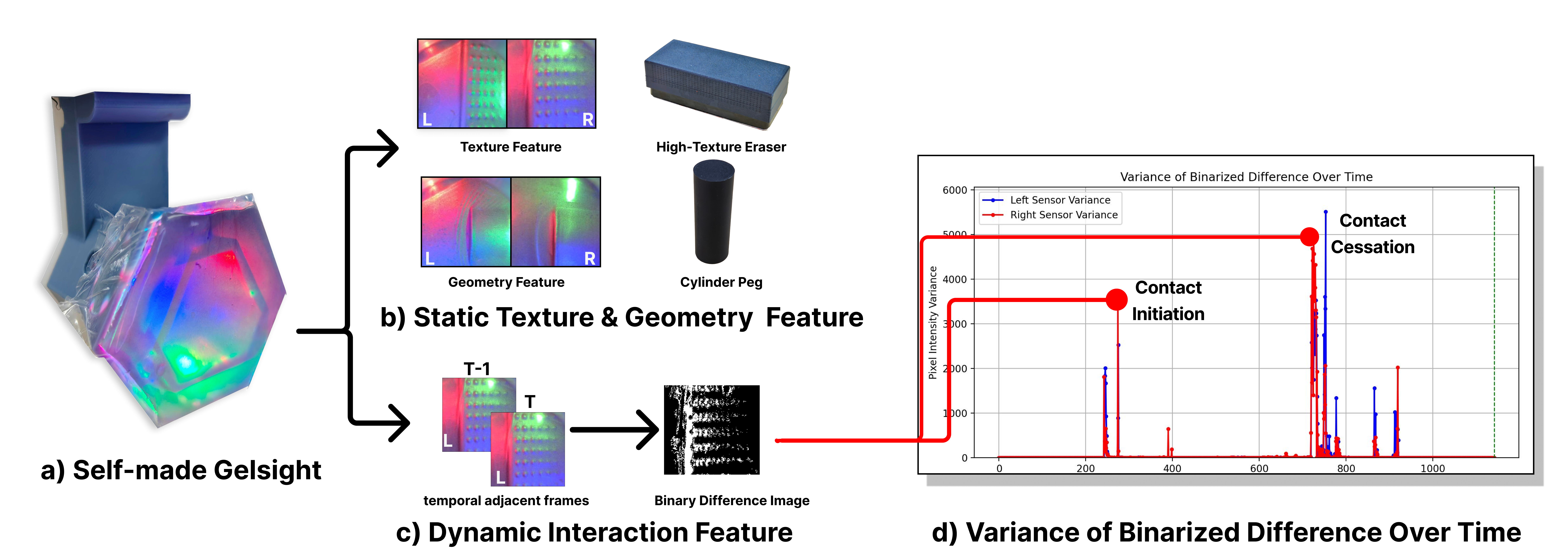}
    \caption{\textbf{Visuotactile sensor dual-channel representation}. 
    (a) The self-made, low-cost Gelsight sensor used for data acquisition. (b) A single static frame provides object properties, such as texture and local geometric profile. (c) Temporal information across frames captures features of dynamic interaction, visualized here via a binary difference image between adjacent frames. (d) Analyzing temporal changes, specifically the variance of the binarized difference over time, explicitly reveals events like contact initiation and cessation, facilitating the learning of policy-relevant features and perception of subtle changes.
    }
    \label{fig:hardware}
    \vspace{-4mm}
\end{figure}

To address the fusion challenge and leverage the rich contact information, we introduce GelFusion, a framework that integrates visuotactile feedback and vision inputs for policy learning in visual-constrained tasks. 
GelFusion employs a modality fusion mechanism based on cross-attention to extract task-related contextual information between visual features and tactile features.
The contextual information serves as complement visual features, particularly when precise contact information is paramount for task success.
To sufficiently extract useful features from visuotactile images, GelFusion proposed a dual-channel visuotactile feature representation (Fig.~\ref{fig:hardware}) which contained two complementary streams explicitly extracted from the tactile modality:

\begin{itemize}[leftmargin=5mm]
\vspace{-2mm}
\item \textbf{High-dimensional spatial features} are extracted via a CNN (ResNet) backbone applied to each individual tactile frame. The features capture fine-grained details about the object's surface texture and the precise geometry of the current contact patch, which is fundamental for understanding static object properties.
\vspace{-1mm}
\item \textbf{Low-dimensional dynamic features} are derived from calculating differences between consecutive tactile frames, specifically encoding crucial temporal interaction events, such as contact initiation or cessation, and subtle pressure changes over time. This compact representation isolates critical mechanical events vital for dynamic control, which might otherwise be obscured within the high-dimensional static frames \cite{zhou2024t}.
\vspace{-2mm}
\end{itemize}

We provide extensive empirical validation across diverse contact-rich tasks, including surface wiping, precise peg insertion, and fragile object pick-and-place. Our results demonstrate that GelFusion outperforms relevant baselines (vision-only and other fusion methods), confirming the effectiveness of the overall framework in enhancing task success rates, especially when nuanced contact control is required.

\section{Related Work}

\textbf{Visuotactile Sensing in Manipulation}
 GelSight, a classic visuotactile sensor, employs an internal camera to capture high-resolution images of contact-induced soft membrane deformation, effectively encoding detailed tactile information \cite{yuan2017gelsight,yuan2016estimating}.
There are mainly two approaches to utilize this rich information.
One explicitly extracts features, such as marker displacement on the membrane corresponding to the changing force field on the silicone surface, for subsequent combination with manipulation actions or encoding into network training \cite{Takahashi2024StableOP,Guo2023EstimatingPO,zhao2024ifem2,xue2025reactive}.
Another approach is to directly encode the images, learning latent patterns, and then integrating these with downstream control policies or tasks \cite{lin2024generalize,yu2023mimictouch,castano2024grasping,Xu2024UniTDE,Zhao2024TransferableTT}.
Inspired by the work of Guo et al. \cite{Guo2023AerialIW} and Zhou et al. \cite{zhou2024t}, we recognize the importance of both aforementioned approaches for representing distinct information levels: static texture and dynamic interaction. Consequently, we designed a dual-channel representation method suitable for marker-less sensors to capture comprehensive contact information.

\textbf{Imitation Learning for Action Generation.}
 Leveraging diffusion model exceptional ability to model complex multimodal distributions and generate diverse, high quality actions, diffusion models are emerging as a promising approach for robotic policy learning \cite{liu2024rdt,ke2024bi3d,mishra2024reorientdiff, mees2024octo, song2025survey}.
Diffusion Policy demonstrates excellent robotic control performance by directly generating diverse action sequences conditioned on observations images \cite{chi2023diffusion}.
A performance bottleneck for Diffusion Policies lies in their reliance on incomplete or occluded observations. Existing methods to improve observations, whether through hardware (UMI \cite{chi2024universal}) or active perception \cite{chuang2024active}, have limited generality in unstructured environments. To address this, we propose a vision-led multimodal fusion framework that supplements vision with rich contact information from visuo-tactile sensing.

\textbf{Sensing fusion for policy learning.}
 Multimodal fusion aims to provide policy learning with a more robust and comprehensive understanding of the states.
Contrastive learning has been applied to visuo-tactile alignment \cite{yang2024binding} and explored for diffusion policy \cite{george2024vital}. However, it is limited by the distribution of pre-training data, making direct generalization to unseen scenarios difficult.
Attention mechanisms are also applicable to cross multimodal fusion
 \cite{li2022see,zhang2023grasp,li2025tvt}, with ManiWav  \cite{liu2024maniwav} applying it to combine audio and visual modalities for the conditioning diffusion policy. However, we observed that directly applying such attention mechanisms to visuo-tactile fusion can potentially degrade the quality of visual observations. Inspired by cross-attention methods \cite{feng2024play}, we propose a vision-led fusion scheme that, by carefully designing the query mechanism, ensures visual feature remains undisturbed during the fusion process.

\section{Method}

\subsection{Policy Design}
\begin{figure}[h]
    \centering
    \includegraphics[width=0.99\textwidth]{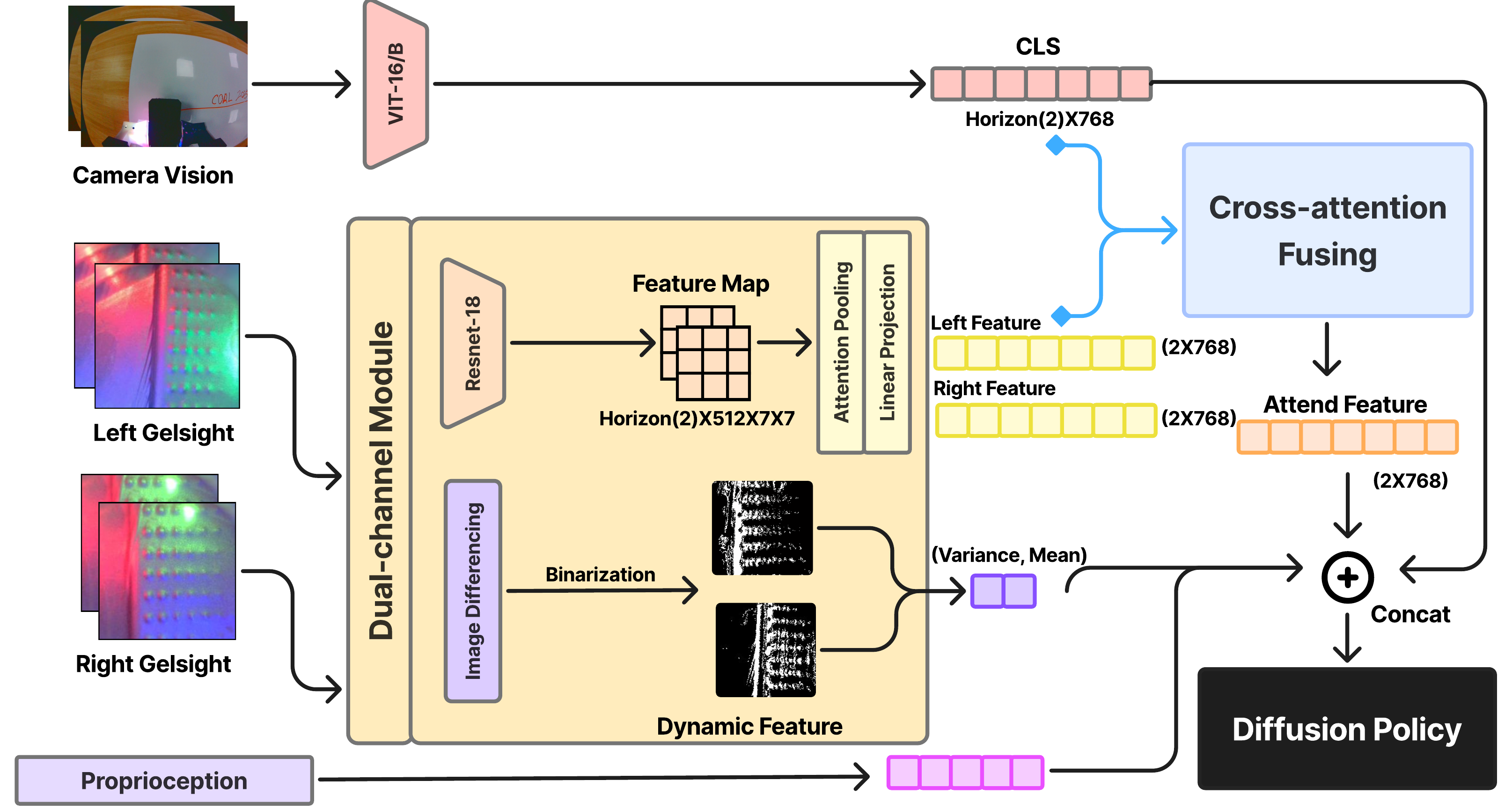}
    \caption{\textbf{Network Architecture.}
    }
    \label{fig:Network}
    \vspace{-4mm}
\end{figure}

We used a standard Diffusion Policy Unet structure network as the framework for the downstream conditional denoising model \cite{chi2023diffusion} (Fig. \ref{fig:Network}).
Using the UMI configuration \cite{chi2024universal}, 2 timesteps are used as observations, and 16 action steps are learned as a whole sequence.
The model outputs a 10-degree-of-freedom action, including end-effector positions, 6D orientation representations, and 1D gripper openness values.

\textbf{Vision Encoder.} We used a CLIP-pretrained ViT-B/16 model \cite{Dosovitskiy2020AnII}, processing 224×224-pixel images in two-frame sequences with random cropping and color jittering. Based on domain expertise, we fine-tuned the model for task adaptation. The classification token served as the visual feature $\mathbf{F}_v$ for multimodal fusion.

\textbf{Tactile Encoder.} Our tactile encoder processes left/right gripper GelSight images independently through a dual-channel architecture, extracting geometric and dynamic features from synchronized 224×224 inputs for contact state representation.

\begin{itemize}[leftmargin=5mm]
    \vspace{-2mm}
    \item \textbf{Geometric Feature Channel} focuses on extracting high-dimensional geometric and textural features. We employ a ResNet-18 encoder \cite{He2015DeepRL} (training from scratch) to process each frame. To obtain a compact one-dimensional feature vector, attention pooling is applied over the spatial dimensions of the 512-channel, 7x7 feature map generated by the encoder. All feature vectors of frames in observation horizon are concatenated and then linearly projected to match the dimensionality of the visual features.
    The images from the left and right GelSight sensors are processed independently to generate the tactile static geometry representations $\mathbf{F}^{l}_{T}$ and $\mathbf{F}^{r}_{T}$.
    \item \textbf{Dynamic Feature Channel}: Complementary to the geometric channel, this channel extracts temporal dynamics.
    The spatiotemporal residual between consecutive tactile frames is binarized using a threshold, prioritizing the encoding of relative change patterns over absolute signal magnitudes.
    Focusing on relative dynamics aims to provide consistent features for policy learning across different conditions.
    By explicitly modeling the occurrence and location of interaction events (contact initiation/cessation, pressure shifts) through these changes, the representation makes it easier for the policy to learn the underlying interaction patterns. Subsequently, the spatial mean and variance are calculated across the binarized residual image to generate a low-dimensional (2D) feature vector $\mathbf{F}_{\text{dyn}}$. The mean reflects the overall proportion of changing area, while the variance indicates the spatial dispersion of these changes, thereby summarizing the interaction's dynamic progression.
    \vspace{-1mm}
\end{itemize}

We treat these features distinctly: the extracted high-dimensional geometric features are fused with visual input, while the low-dimensional dynamic features undergo separate processing within our feature fusion module (Fig.~\ref{fig:Network}).

\textbf{Sensory Fusion.} Our cross-modal fusion strategy employs a cross-attention mechanism to integrate information effectively. Attention weights, denoted as $[W_V, W_{T}^{l}, W_{T}^{r}]$, are computed to quantify the relevance of each modality's feature representation with respect to the visual query feature $\mathbf{F}_v$. These weights are derived from the scaled dot-product similarity between the visual query $\mathbf{Q}$ (derived from $\mathbf{F}_v$) and the key matrix $\mathbf{K}$ (constructed from all modal features: visual $\mathbf{F}_v$, left visuo-tactile $\mathbf{F}_{T}^{l}$, and right visuo-tactile $\mathbf{F}_{T}^{r}$), followed by a softmax normalization. Each weight ($W_V, W_{T}^{l}, W_{T}^{r}$) thus reflects the normalized importance assigned to the corresponding modality (visual, left tactile, right tactile) based on the visual query's context.

    \begin{wrapfigure}{r}{0.5\textwidth}
\vspace{-4mm}
\includegraphics[width=0.5 \textwidth]{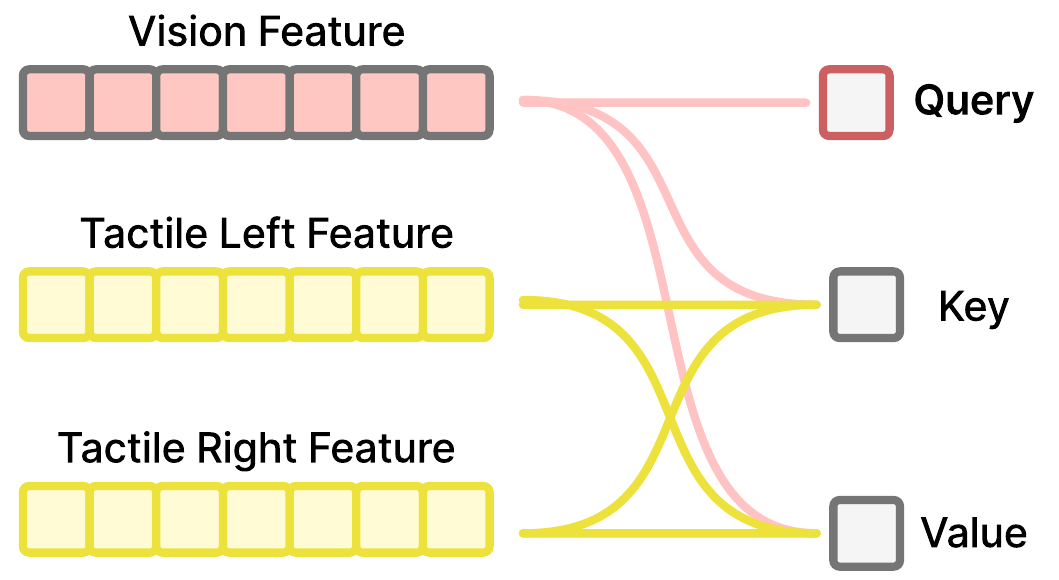}
    \caption{\textbf{Vision-led Cross-attention Fusion.} }
    \vspace{-2mm}
    \label{fig:model}
\end{wrapfigure}

These attention weights are then used to compute an attended feature representation, $\mathbf{F}_{\text{att}}$, by performing a weighted summation of the value features corresponding to each modality:
$$
\mathbf{F}_{\text{att}} = W_V \mathbf{F}_V + W_{T}^{l} \mathbf{F}_{T}^{l} + W_{T}^{r} \mathbf{F}_{T}^{r}
$$
 This weighted sum $\mathbf{F}_{\text{att}} \in \mathbb{R}^{B \times D}$ effectively aggregates information across modalities, dynamically emphasizing those features deemed most relevant by the attention mechanism relative to the visual query (Fig. \ref{fig:model}).

Concatenating the original visual feature $\mathbf{F}_v$ and the attended feature $\mathbf{F}_{\text{att}}$ yields the final high-dimensional feature $\mathbf{F}_{\text{fusion}}$, which retains original visual information while incorporating cross-modal attention context. This feature $\mathbf{F}_{\text{fusion}}$ is then concatenated with low-dimensional features $\mathbf{F}_{\text{dyn}}$ and proprioception features to serve as the condition guide diffusion policy denoise.

\section{Experiments}
\vspace{-2mm}

We evaluate GelFusion on three contact-rich manipulation tasks: precise stick insertion, delicate fragile object picking, and dynamic surface wiping. These tasks were chosen as they inherently challenge vision-only approaches due to complex contact dynamics and potential visual limitations.

\textbf{Hardware set.}
Data collection was performed via teleoperation directly on the Fairino FR5 robotic arm, providing direct access to accurate proprioceptive data like joint position and velocity. Visual input was exclusively sourced from a wrist-mounted camera. Additionally, a self-made visual-tactile GelSight sensor was integrated into the system.

\subsection{Surface Wiping Task}

\begin{figure}[t]
    \centering
    \includegraphics[width=\textwidth]{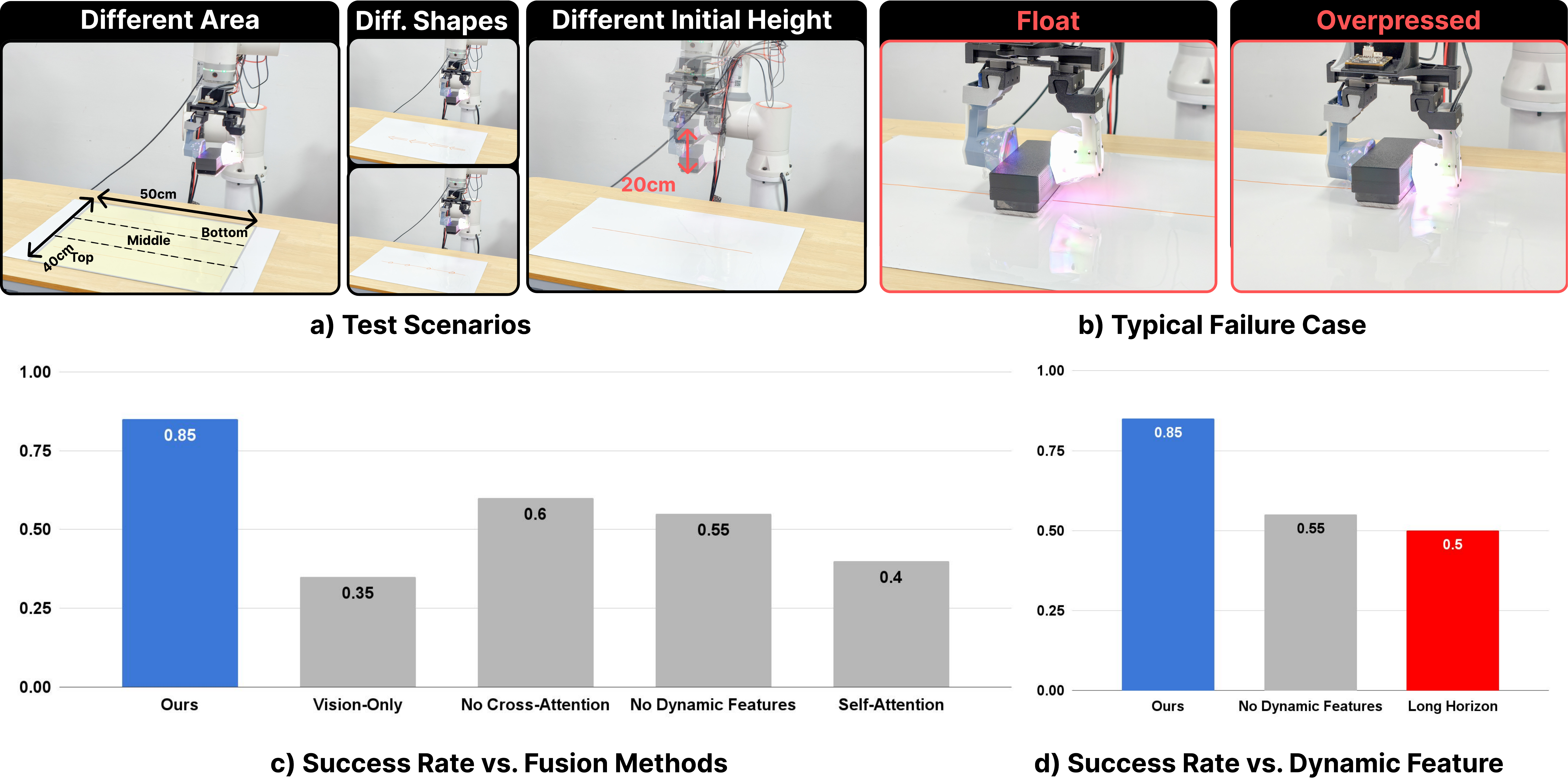}
    \caption{\textbf{Wiping Evaluation.} (a) Test scenarios used to evaluate policy generalization, including variations in wiping area, line shape, and initial height compared to training conditions. (b) Typical failure cases observed: the eraser floating above the table (Float) and the gripper overpressing the eraser (Overpressed). (c) and (d) show experimental results, comparing the success rate of our method ('Ours') against ablation baselines and variants.}
    \label{fig:wipe}
    \vspace{-4mm}
\end{figure}

This experiment evaluates how GelFusion's visuotactile feedback (utilizing its dual-channel tactile representation) assists robots in the surface wiping task by overcoming the limitations of vision in establishing precise contact and maintaining consistent pressure/posture (keeping the eraser flush) (e.g., depth ambiguity, reflections), thereby achieving effective surface cleaning.

\textbf{Test Scenarios:} As shown in Fig. \ref{fig:wipe}(a), we conducted 20 inference tests for policies of different architectures. Among them, 15 tests involved wiping straight lines of approximately 40cm in length at random starting positions within the "Top," "Middle," and "Bottom" sections of the table, while the remaining 5 tests evaluated the performance of random line shapes at arbitrary positions. The initial height was varied for each test to examine whether the model had overfitted to certain trajectory data.

\textbf{Evaluation Protocol and Baselines:}

\begin{enumerate}[leftmargin=5mm]
    \vspace{-2mm}
    \item \textbf{Full GelFusion:} Our proposed architecture using cross-attention fusion and the dual-channel (ResNet spatial + dynamic difference) tactile representation.
    \vspace{-1mm}
    \item \textbf{Vision-Only Baseline:} The standard Diffusion Policy baseline~\cite{chi2023diffusion} using only wrist-camera visual input, representing the performance without any tactile feedback.
    \vspace{-1mm}
    \item \textbf{No Dynamic Features (CNN Spatial Only):} Removed the temporal difference channel to assess the impact of dynamic tactile cues on capturing contact state changes.
    \vspace{-1mm}
    \item \textbf{No Cross-Attention (Concat Fusion):} Replaced cross-attention with simple feature vector concatenation to evaluate the effectiveness of the attention mechanism for fusion.
    \vspace{-1mm}
    \item \textbf{Self-Attention:} Following Li et al. \cite{li2022see}and Liu et al. \cite{liu2024maniwav}, leveraging mutual interactions across modalities effectively enhances the fusion of visual and other features. This approach should be superior to feature fusion via vector concatenation.
    \vspace{-1mm}
    \item \textbf{Long Horizon:} To compare with explicit methods for extracting interaction dynamic features, a set of carefully designed control groups was additionally included, utilizing a longer GelSight image horizon (2 to 4 frames) to thoroughly explore the correspondence between latent patterns in image variations and states.
    \vspace{-2mm}
\end{enumerate}

\textbf{Findings:}
The quantitative results comparing different methods are presented in Fig. \ref{fig:wipe}(c) and Fig. \ref{fig:wipe}(d): one group (Fig. \ref{fig:wipe}(c)) shows the ablation-based experimental results, and the other group (Fig. \ref{fig:wipe}(d)) compares different methods applied to the key Module / dynamic features.

1) \textbf{Ablation Study Insights:} Our ablation study confirmed the effectiveness of GelFusion's components. The full model achieved the highest success rate. Removing cross-attention (\textit{No Cross-Attention}) led to frequent 'air wiping' (loss of contact, see 'Float' example in Fig. \ref{fig:wipe}(b)), demonstrating the mechanism's importance for robust tactile integration. Removing dynamic features (\textit{No Dynamic Features}) resulted in failures due to excessive force (see 'Overpressed' example in Fig. \ref{fig:wipe}(b)), highlighting this channel's role in modulating contact interactions. Crucially, both ablated models still outperformed the \textit{Vision-Only} baseline, confirming the necessity of tactile feedback for this contact-rich task and validating the specific contributions of our cross-attention fusion and dual-channel tactile representation.

2) \textbf{Fusion Methods Comparison:} Experimental results showed that self-attention fusion had lower accuracy in locating wiping start points than a vision-only policy and generalized poorly to varied start positions, appearing reliant on proprioception. It also failed to effectively use GelSight contact features, often leading to overpressing. Potential reasons include the limited dataset size (even 100 demonstrations are insufficient for contact learning) and the sparse nature of GelSight data generating redundant inputs that impede fusion. Therefore, we carefully designed the visual component in GelFusion to prevent these problems.

3) \textbf{Long Horizon Comparison:} We do not find a better understanding of contact modality features with longer horizons policy. Failures still mostly occur due to overpressing during contact. Perhaps a longer contextual horizon, using architectures like LSTMs \cite{Kapturowski2018RecurrentER} to better capture temporal dependencies could yield better results.

\subsection{Insertion Task}
The peg insertion task evaluates precise alignment capabilities. Its primary challenge lies in handling the inherent visual occlusion during the final insertion phase, mirroring vision-limited scenarios where tactile feedback becomes crucial for success.

\begin{figure}[t]
    \centering
    \includegraphics[width=\textwidth]{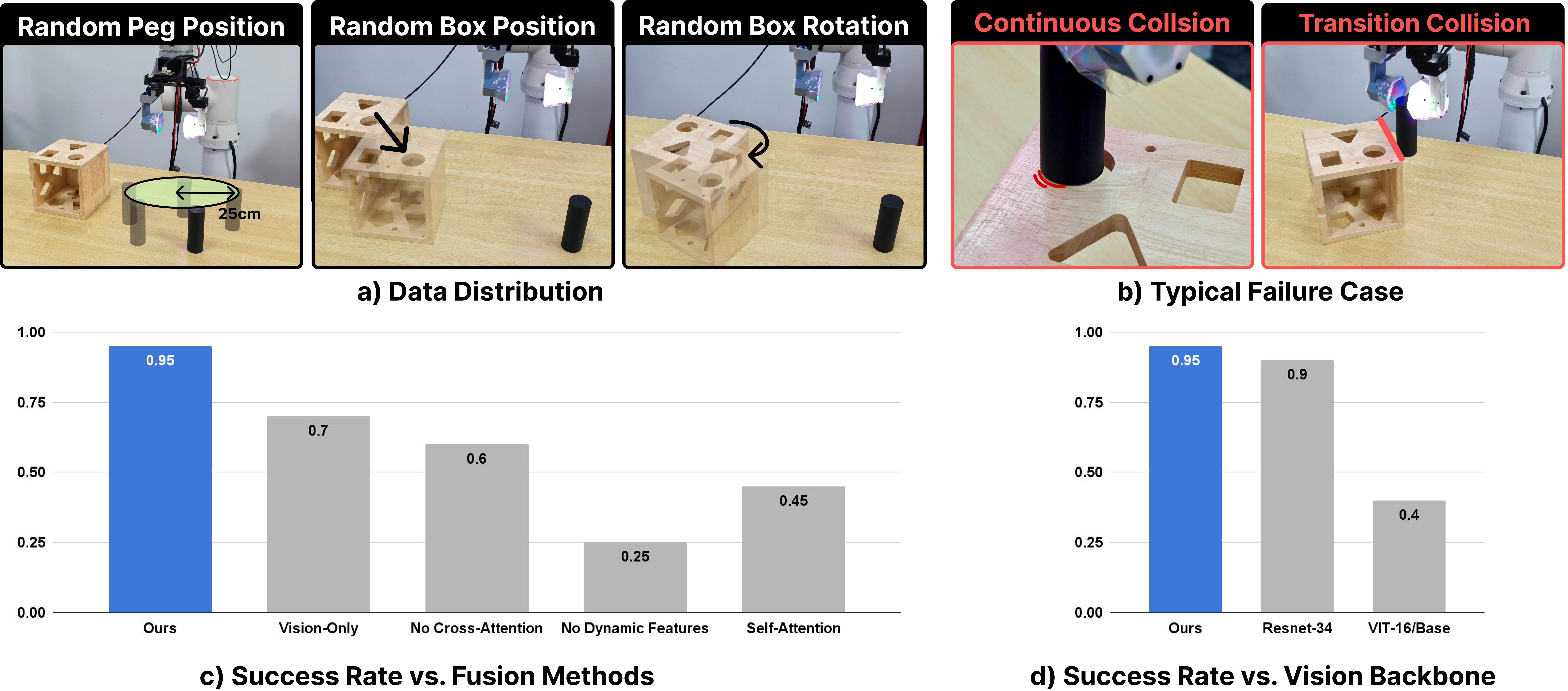}
    \caption{\textbf{Insertation Evaluation.} (a) Randomized evaluation scenarios showing variations in initial peg position, puzzle box position, and puzzle box rotation. (b) Typical failure cases observed: 'Continuous Collision' during insertion and 'Transition Collision' upon approach. (c) Success rate comparison of our method ('Ours') against ablation studies on fusion methods and attention mechanisms. (d) Success rate comparison using different vision backbone architectures.}
    \label{fig:insertation}
    \vspace{-5mm}
\end{figure}

\textbf{Test Scenarios:} Similarly, we conducted 20 inference tests for each model, with randomized variations in initial peg position, puzzle box position, and puzzle box rotation(Fig. \ref{fig:insertation}(a)). For instance, the peg was placed at 60° angles to the left, right, and directly in front, while the box was rotated in 90° increments. These combinations resulted in approximately 12 fixed configurations. The remaining 8 tests were performed using completely random arrangements within the data distribution.

\textbf{Evaluation Protocol and Baselines:} For the ablation experiment setup, we adopted the same four structures as mentioned earlier.
To investigate the influence of the tactile feature extractor's architecture, we also trained and evaluated variants replacing GelFusion's default tactile CNN (presumably ResNet18 based on your profile description):
     \textit{a) ResNet34 Backbone:} Used a deeper ResNet model (ResNet34) for spatial tactile feature extraction.
    \textit{b) ViT Backbone (UMI-Style):} Employed a Vision Transformer (ViT-B/16), fine-tuned similarly to methodologies like UMI, using only the [CLS] token output as the tactile feature representation for fusion.

\textbf{Findings:}Quantitative results for the peg insertion task are summarized in Fig. \ref{fig:insertation}(c) and Fig. \ref{fig:insertation}(d).

1) \textbf{Ablation Study Insights:}
The full GelFusion framework achieved high success rates across challenging peg insertion configurations While the vision-only policy showed effective localization and could complete task phases not requiring fine contact adjustments, it frequently failed during precise interactions due to inherent limitations in visual perception. Key failure modes included inaccurate grasps, and critically, misjudgment of the alignment state due to visual occlusion, which caused excessive force application and catastrophic failures, including sensor damage (typical failures like 'Continuous Collision' and 'Transition Collision' are shown in Fig. \ref{fig:insertation}(b)). These outcomes illustrate the insufficiency of relying solely on vision for contact-rich insertion tasks, even when adequate for simpler sub-steps.

Ablating key components of GelFusion further validated our design. Removing the dynamic interaction channel improved grasp stability over the vision-only baseline via static tactile cues, but still suffered failures from excessive force during alignment, indicating the necessity of dynamic features for reacting to contact forces.
Removing the cross-attention fusion mechanism resulted in failures similar to the vision-only baseline , suggesting its importance for integrating early tactile feedback to refine grasp execution. The study confirms that both the dual-channel tactile representation and the cross-attention fusion mechanism are integral to GelFusion's robust performance in the precision insertion task, overcoming limitations of vision-only approaches.

2) \textbf{Tactile Backbone Comparison:}
Evaluation of alternative tactile feature extractors (backbones) showed that a ResNet18 trained from scratch yielded the best task performance. Conversely, fine-tuning a pre-trained ViT-B/16 using only the [CLS] token (UMI-style) resulted in significantly worse performance, with failure modes often resembling the vision-only baseline. This suggests potential challenges with information compression via the [CLS] token or the domain gap between visual pre-training and tactile sensing. Using a deeper ResNet34 backbone offered no significant benefit over ResNet18, indicating that ResNet18's representational capacity was adequate for encoding the necessary tactile features for this task.

\subsection{Chips Pick Task}

The task objective is robotic grasping of highly fragile objects like potato chips. As brittle materials prone to sudden fracture under excessive force, precise force control is crucial for maintaining their structural integrity. This task highlights the limitations of relying solely on wrist-camera vision, which lacks the necessary force-sensitive feedback to prevent such instantaneous damage.

\begin{figure}[t]
    \centering
    \includegraphics[width=\textwidth]{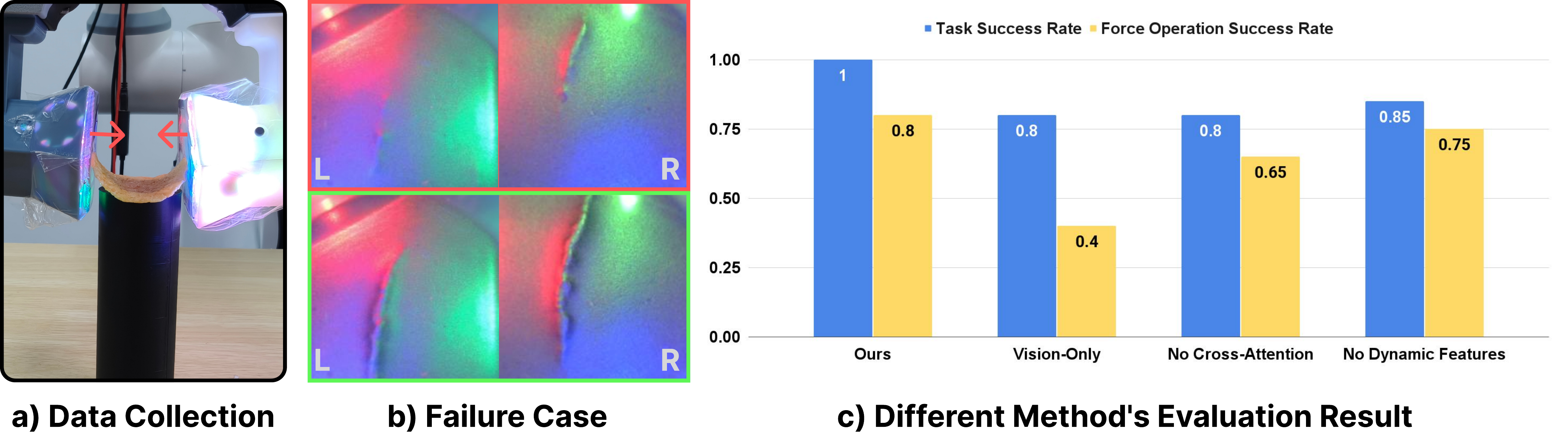}
    \caption{\textbf{ChipPick Evaluation.} (a) demonstrates the method of picking up potato chips, which highly tests the control of force; (b) shows common failure cases (marked in red) when only visual features are input, where the policy always chooses to grasp the chips too lightly; (c) to address the issue in (b), we had to reassess the success rate labeling, shifting focus from task completion to fine-grained operational details.}
    \label{fig:chipPick}
    \vspace{-6mm}
\end{figure}

\textbf{Data collection:}
A dataset of 50 trajectories was collected for grasping intact, saddle-shaped potato chips(Fig. \ref{fig:chipPick}(a)). The grasping strategy involved gripping the chip ends. Demonstrations  achieve successful grasps without fracture; failed attempts were excluded.

\textbf{Test Scenarios:}
For this task, we conducted ablation studies evaluating different framework configurations, including those incorporating tactile feedback and the vision-only baselines.  Each configuration was evaluated over 10 trials to comprehensively assess its performance on fragile object grasping.

\textbf{Findings:}
Experimental results indicated that the vision-only policy, unlike in the wiping task, tended towards very gentle grasps for chip manipulation, avoiding excessive force. Although gentle grasps sometimes achieved transport, the primary evaluation focused on the policy's ability to replicate demonstrated force-controlled grasps. Therefore, grasps were considered failures for force control analysis if they were either too forceful (causing breakage) or too gentle (insecure or significantly deviating from demonstration force levels, see Fig. \ref{fig:chipPick}(b)). Evaluation involved quantifying relative grasp forces via differential imaging and comparing the policy's applied force distribution against that of successful demonstrations.

Under this refined success criterion, experiments incorporating tactile feedback generally demonstrated superior performance compared to the vision-only approach (Fig. \ref{fig:chipPick}(c)). However, within the tactile-enhanced configurations, the performance differences between various tactile information modules were less pronounced. This could potentially be attributed to the protective properties of the silicone gripper material itself, which may have mitigated some of the inherent challenges associated with grasping brittle objects, regardless of the specific tactile features utilized.

\section{Conclusion}
In conclusion, we presented GelFusion, a framework designed to enhance policy learning (like Diffusion Policies) for operation in visually limited conditions by integrating rich visuotactile feedback. Our approach, utilizing a dual-channel feature representation capturing contact details and fused via cross-attention, demonstrates how tactile data can potentially compensate when visual information is compromised. Evaluations on contact-rich tasks showed performance improvements compared to baselines, particularly in these challenging scenarios. By addressing contact-richness and visual limitations simultaneously, this work contributes a step towards enabling more robust and reliable robotic manipulation in challenging in-the-wild settings.

\clearpage
\acknowledgments{The authors would like to thank Haosen Yang and Zhao Tang for their discussions and feasibility analysis. We would like to thank all Gair Lab members: Jing Wang, Hairong Qu, et al. Thanks are also extended to Lili Ma and Xin Cheng for providing emotional support.}

\bibliography{main}

\vspace{4mm}
{\Large \textbf{Limitation}}

Despite the effectiveness demonstrated by GelFusion in contact-rich tasks under visual constraints, this study has several limitations.

Firstly, the dataset size is relatively limited, for all tasks, we only used 50 demostrations for training. While the collected demonstrations were sufficient to validate the framework's feasibility, they may not fully capture the extensive diversity of contact variations and complex scenarios encountered in real-world applications. This could potentially restrict the generalization robustness of the learned policy when facing novel or out-of-distribution situations. Furthermore, this limited data scale might also contribute to the suboptimal performance observed in some baseline methods, particularly those attempting complex cross-modal fusion like self-attention. As suggested by our findings , such approaches may require significantly more data to effectively learn intricate visuotactile relationships and overcome challenges posed by potentially sparse or redundant tactile inputs. Expanding the dataset's scale and diversity remains an important direction for future improvement, both for our method and for fairly evaluating complex fusion techniques.

Secondly, the method for processing dynamic tactile information could be enhanced. Currently, we capture dynamic interaction events by calculating differences between consecutive tactile frames and extracting statistical features ($\mathbf{F}_{dyn}$). Although concise and shown to be beneficial, this approach might simplify the rich temporal dynamics inherent in the contact process. Potentially valuable information, such as subtle rates of change in pressure or shear force patterns, might not be fully utilized. Exploring more sophisticated temporal modeling approaches (e.g., RNNs, LSTMs, or TCNs) could allow for a more comprehensive representation and leveraging of dynamic tactile features, potentially offering further performance gains, particularly in tasks requiring highly nuanced force control or rapid dynamic responses.

\newpage
\appendix
{\Large \textbf{Appendix}}
\section{Hardware Details}
\subsection{Self-made Gelsight}

\begin{wrapfigure}{r}{0.5\textwidth}
\vspace{-4mm}
\includegraphics[width=0.5 \textwidth]{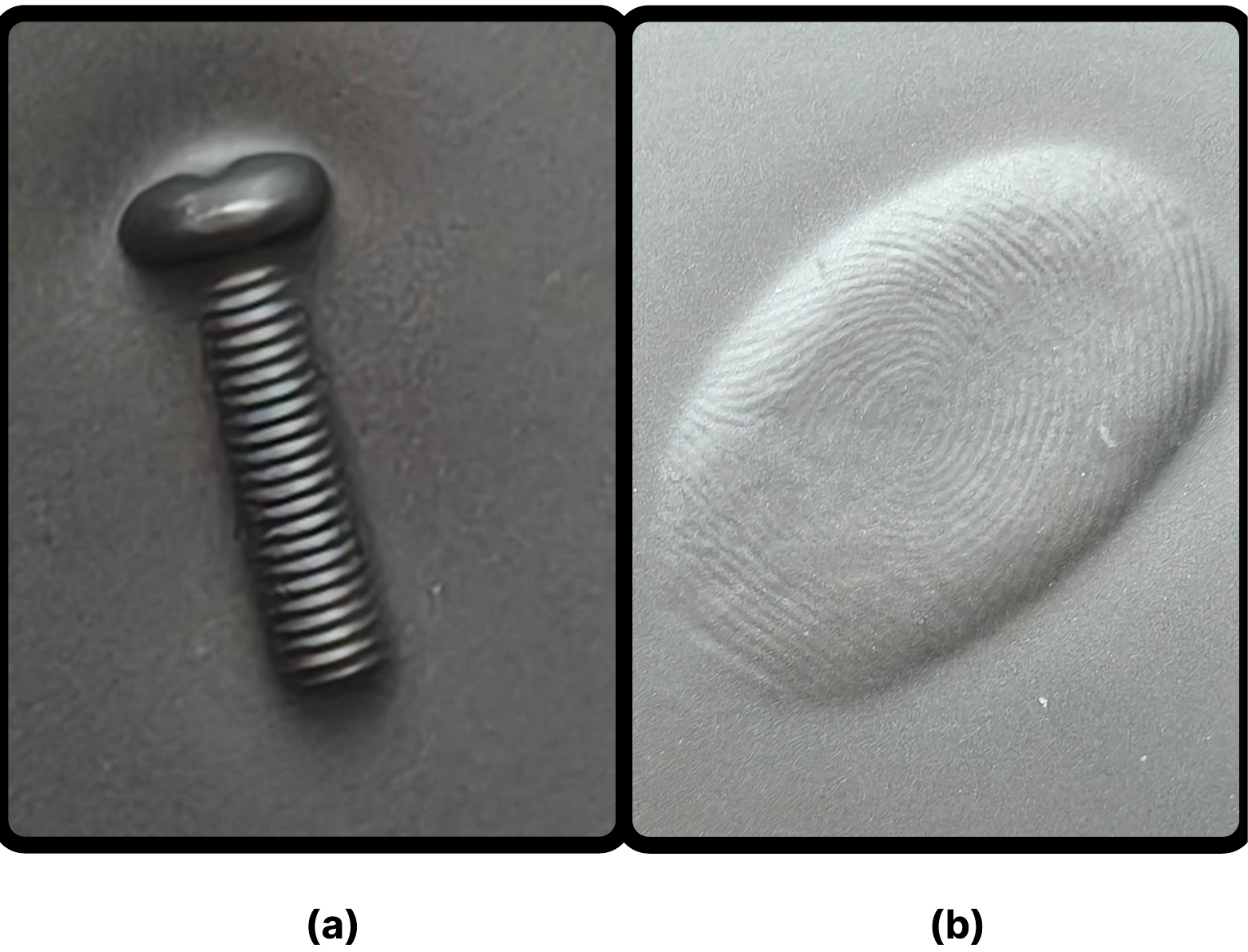}
    \caption{\textbf{Silicone Layer}: (a) M3 screw is pressed on silicone layer. (b) Clear fingerprint.}
    \vspace{-5mm}
    \label{fig:silicone}
\end{wrapfigure}

Rich tactile information is acquired by a self-made GelSight sensor integrated into the gripper. The rationale for this self-made approach centers on achieving a low-cost, scalable, and readily deployable sensor configuration, thereby expanding application potential, and on simplifying the fabrication of the damage-prone silicone component to alleviate testing constraints. Despite the low sensor assembly cost of approximately 30 USD, largely attributed to the camera, good surface precision sensing is maintained.

Silicone fabrication is simplified, requiring no complex vacuum or pressure equipment through the use of a high-transparency, self-deaerating two-part addition-cure silicone. As shown in Figure \ref{fig:silicone}, pressing an M3 screw onto the silicone layer leaves a clear fingerprint, demonstrating the sensor's ability to capture fine details.

The reflective surface layer is prepared by incorporating aluminum silver powder and applying it thinly onto the substrate, referencing established methods~\cite{yuan2017gelsight,Taylor2021GelSlim3H}. Curing is accelerated by maintaining the 3D-printed base at 35°C, enabling a complete fabrication cycle in approximately 5 hours, with the capacity for concurrent production of multiple units.

\begin{wrapfigure}{r}{0.33\textwidth}
\vspace{-4mm}
\includegraphics[width=0.33 \textwidth]{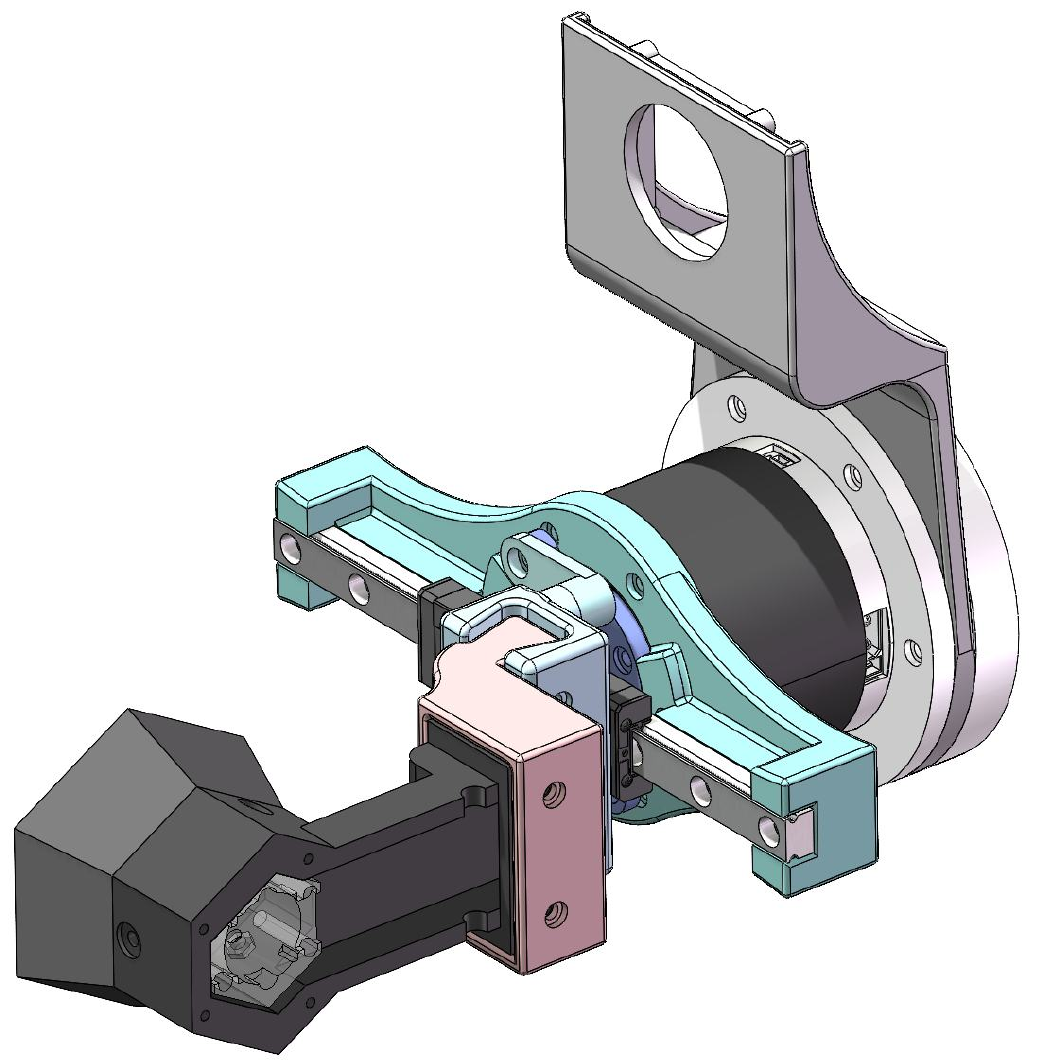}
    \caption{Quick-release Gripper }
    \vspace{-10mm}
    \label{fig:gripper}
\end{wrapfigure}
To enhance versatility and ease of use, a quick-release mechanism was developed for the custom-built GelSight sensor and its gripper attachment, enabling rapid transition between robotic arm-based testing and field data collection leveraging the UMI architecture. The quick-release gripper is shown in Figure \ref{fig:gripper}.

\subsection{Hardware Configuration}
As detailed in Section 4, the experimental setup employs a Fairino FR5 robotic arm operating at 100 Hz. Sensory data is captured at 30 Hz from a main-view camera and both GelSight sensors, with all timestamps synchronized to the main-view camera. Model input for inference consists of two timesteps of historical observations. The main-view camera is a modular UVC with a 150-degree wide-angle lens, and the GelSight sensors use Logitech C270 webcams focused on the reflective surfaces. Inference is performed on an Nvidia 4060 Ti under Ubuntu 20.04 at 2 Hz.

\section{Model Details}

In this work, for each designed task, the entire model underwent end-to-end training, conducted on 2 NVIDIA A800 GPUs for approximately 50 epochs with a batch size of 64. Training utilized BF16 mixed precision for faster convergence (roughly double FP32 speed) with negligible impact on inference. Exponential Moving Average (EMA) was applied to the weights during this process. Optimization was performed using AdamW (betas=[0.95, 0.999], eps=1.0e-8, weight decay=1.0e-6)~\cite{chi2023diffusion}. The initial learning rate for this training was set to 3e-4. Fine-tuning the released pre-trained Vision Transformer (ViT) model will employ a learning rate of approximately 3e-5.

\textbf{Self-Attention Baseline Implementation Details.} The Self-Attention baseline employs a standard Transformer encoder layer for multimodal fusion~\cite{li2022see, liu2024maniwav}. It takes the feature vectors from each modality as a sequence input. The Transformer layer processes this sequence to learn interactions between modalities, outputting a fused multimodal representation for the policy.

\section{Task Details}

\begin{figure}[h]
    \centering
    \includegraphics[width=\textwidth]{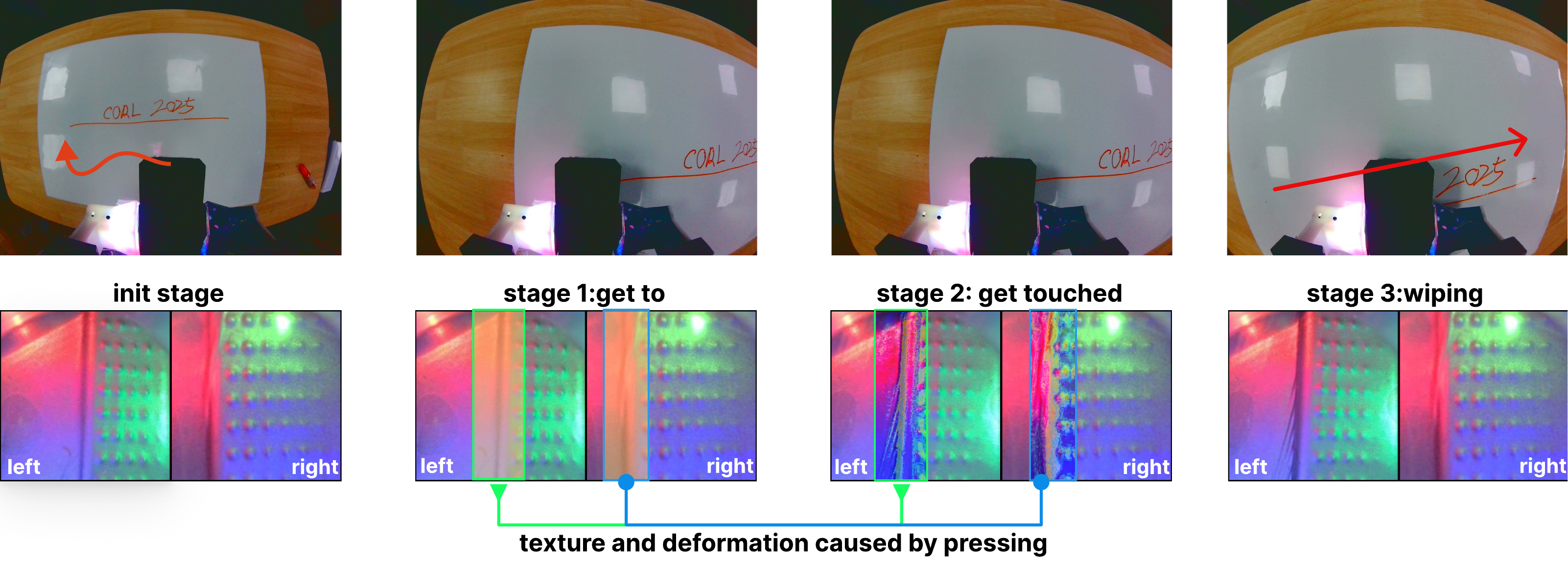}
    \caption{\textbf{Pipeline of Wiping Task}}
    \label{fig:wiping_pipeline}
    \vspace{-8mm}
\end{figure}
\subsection{Wiping Task}

\textbf{Experimental Setup}

It has two critical contact phases: Stage 1: establishing contact from an arbitrary height and Stage 2: maintaining consistent pressure/orientation while wiping. To highlight failure modes, the wiping action is performed only once in a continuous left-to-right pass. The pipeline for the wiping task is illustrated in Figure \ref{fig:wiping_pipeline}.

The experimental setup focuses on the wiping action itself; the robot begins with the eraser already grasped in a predefined pose. The custom GelSight sensor employed has a sensing area smaller than the eraser's contact surface, implicitly requiring the policy to manage relative motion to keep key contact features within view.

\begin{wraptable}{r}{0.5\textwidth}
\vspace{-10pt}
\centering
\begin{tabular}{ccc}
\hline
\textbf{Line Type} & \textbf{Position} & \textbf{Num.} \\ \hline
Line               & Top               & 10            \\
Line               & Middle            & 10            \\
Line               & Bottom            & 10            \\
Arrow              & Top               & 5             \\
Arrow              & Middle            & 5             \\
Arrow              & Bottom            & 5             \\
"CORL 2025"        & Middle            & 5             \\ \hline
\end{tabular}
\caption{Distribution of Wiping Task.}
\label{tab:wiping_demonstrations}
\vspace{-8mm}
\end{wraptable}

\textbf{Data Collection Details}

We collected 50 demonstrations on a 60cm x 42cm whiteboard. To promote policy generalization across the workspace, demonstrations were distributed across three notional horizontal regions (top, middle, bottom). While the majority consisted of wiping straight lines (~40cm), we introduced variability by including demonstrations of hand-drawn patterns (arrows, "CORL 2025"). This hand-drawn nature inherently added slight variations to the trajectories and target shapes within the dataset. All demonstrated motions were continuous left-to-right passes. The distribution of these demonstrations is detailed in Table \ref{tab:wiping_demonstrations}.

\textbf{Methodology for Assessing Task Success}

Task success is assessed based on two main failure types: \textbf{floating} and \textbf{overpressure}. Floating fails if the residual un-erased line is over 10 cm. Overpressure fails if the GelSight silicone is damaged or if the eraser boundary slides out of the GelSight camera's view due to the relative sliding motion generated during contact.

\subsection{Insertation Task}

\begin{figure}[h]
    \centering
    \includegraphics[width=\textwidth]{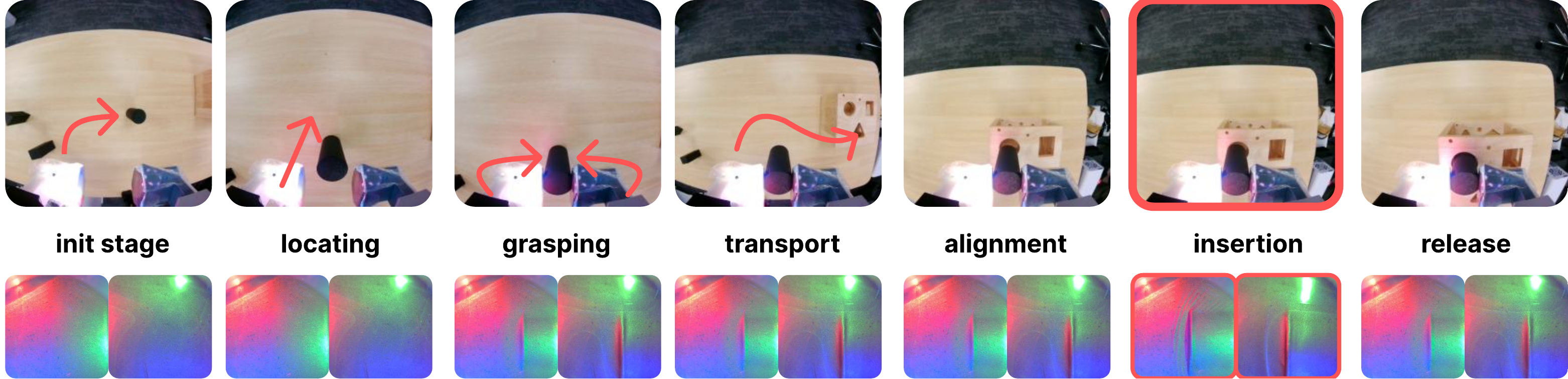}
    \caption{\textbf{Pipeline of Insertion Task}}
    \label{fig:insertion_pipeline}
    \vspace{-6mm}
\end{figure}

\textbf{Experimental Setup}

The insertion task encompasses a sequence of critical phases: grasping, transport, alignment, insertion, and release. To assess the robustness of the policy, significant randomization was introduced across several key parameters. The pipeline for the insertion task is shown in Figure \ref{fig:insertion_pipeline}.

The position of the peg was randomly sampled within a 25cm radius circle relative to the robot's base on the workspace. The pose of the puzzle box was also randomly sampled within a reachable workspace area, with its orientation systematically varied. Specifically, demonstrations were collected covering a range of puzzle box orientations, including fixed rotations (0°, 90°, 180°, and 270° relative to a nominal forward-facing orientation) and fully random angles, to ensure the policy learned to handle diverse visual presentations. Furthermore, the presence of square and triangular holes on the puzzle box surface, alongside the target circular hole, served as distractors, necessitating the policy's ability to accurately identify the target based on visual input.

\textbf{Data Collection Details}

A total of 50 successful demonstrations were collected for policy learning. These demonstrations reflected a core insertion strategy that prioritized visual alignment, where the policy first attempts to align the peg with the hole using visual feedback ("coarse alignment"). Tactile feedback was then utilized for "fine alignment" adjustments if contact occurred before successful insertion.

\begin{table}[h]
\centering
\begin{tabular}{cccl}
\hline
\textbf{Rotation Angle (Deg.)} & \textbf{Transition X (cm)} & \textbf{Transition Y (cm)} & \multicolumn{1}{c}{\textbf{Num.}} \\ \hline
0   & -25      & -25      & 10 \\
90  & 25       & -25      & 10 \\
180 & -25      & 25       & 10 \\
270 & 25       & 25       & 10 \\
Any & $<\pm25$ & $<\pm25$ & 10 \\ \hline
\end{tabular}
\vspace{2mm}
\caption{Distribution of Insertion Task Demonstrations by Puzzle Box Pose.}
\label{tab:insertion_demonstrations}
\vspace{-2mm}
\end{table}

The data distribution captured this strategy: approximately 67\% of demonstrations achieved insertion through visual alignment, with 33\% requiring contact-triggered adjustments. The specific distribution of puzzle box poses used in the demonstrations is detailed in Table \ref{tab:insertion_demonstrations}.

\textbf{Methodology for Assessing Task Success}
This multi-stage task requires a high success rate, with completion defined by success in all phases. Actual success rates are approximately 20\% higher than anticipated, attributed to a visually-guided data distribution and the Diffusion Policy baseline's generalization. A comprehensive analysis of all cases is performed to demonstrate the framework's contribution over the baseline. Notably, forceful insertions succeeding despite needing adjustment are classified as policy failures, as they are inconsistent with the intended strategy and data distribution.

\subsection{Chips Pick Task}

\begin{wrapfigure}{r}{0.5\textwidth}
\vspace{-4mm}
\includegraphics[width=0.5 \textwidth]{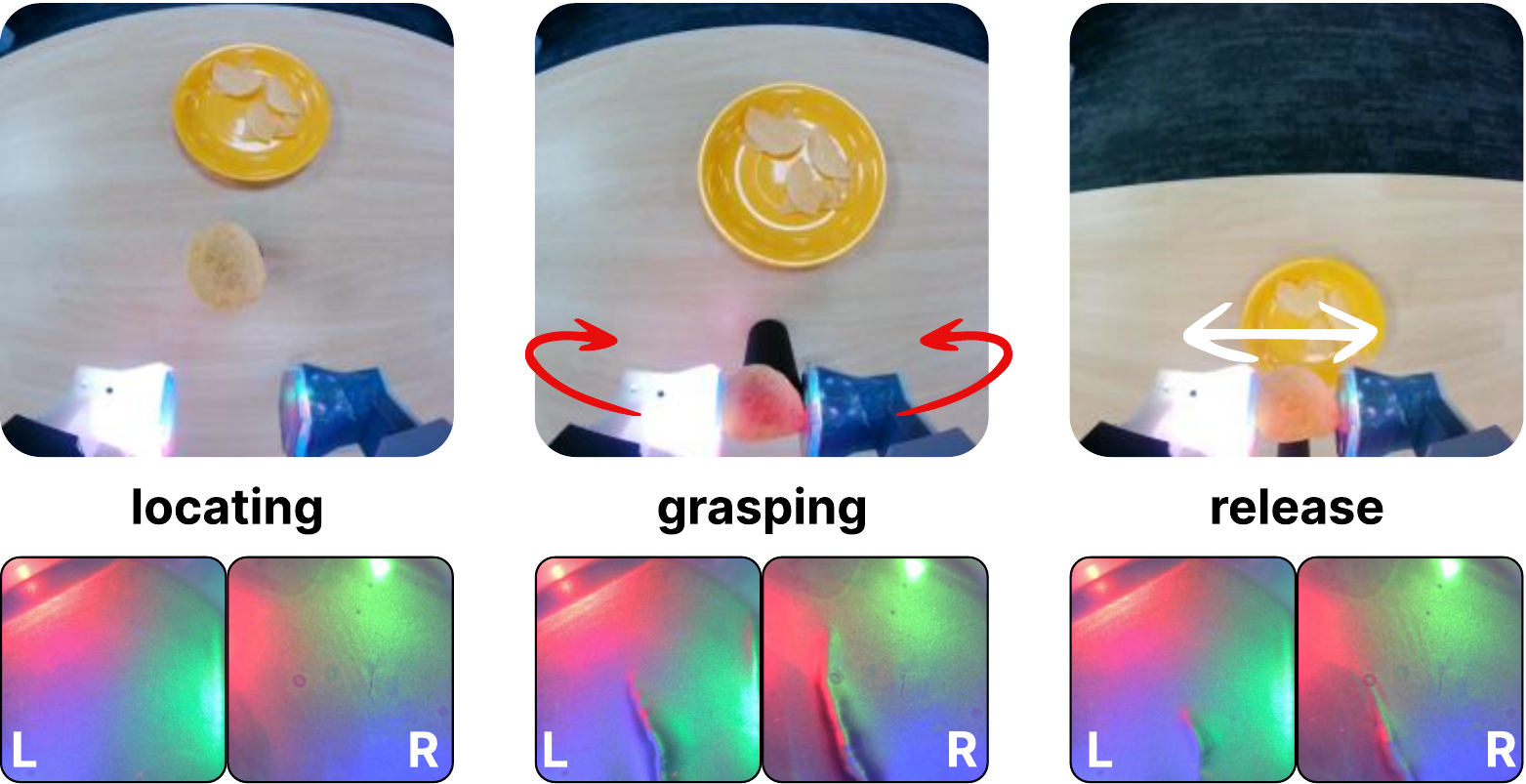}
    \caption{\textbf{Pipeline of Chips Pick Task}}
    \vspace{-4mm}
    \label{fig:chips_pick_pipeline}
\end{wrapfigure}

\textbf{Experimental Setup}
For this task, which is designed with relative simplicity and primarily focuses on the chip grasping process, chips are positioned at a fixed location and orientation relative to the gripper's initial stance for subsequent retrieval; following successful grasping, the chip is then transported above a designated plate and released, thereby depositing it into the container. The pipeline for the chips pick task is shown in Figure \ref{fig:chips_pick_pipeline}.

\textbf{Data Collection Details}
A total of 50 grasping examples are likewise collected, with each grasping instance carefully executed at a force level approaching the threshold of chip breakage; the relative magnitude of force in each instance is subsequently characterized by computing the average value of the binary residual image obtained after background subtraction from the GelSight data, and a concerted effort is made to maintain the applied force as consistent as possible across all demonstrations.

\textbf{Methodology for Assessing Task Success}
Throughout the entire process, changes in the GelSight data are meticulously recorded, and the moment of maximal deformation is selected for the calculation of the relative force magnitude; this calculated force is then compared against the data collected during the demonstrations, and if the discrepancy exceeds 20\%, the inference for that particular instance is deemed to have failed.

\end{document}